\documentclass{lncs}

\input{macros.sty}
\usepackage{amsmath}
\usepackage{amssymb}
\usepackage{multirow}
\usepackage{url}
\usepackage{paralist}
\usepackage{graphicx}
\usepackage{rotating}
\usepackage{subcaption}
\usepackage[utf8x]{inputenc}

\usepackage{framed}
\usepackage{color}
\usepackage{enumerate}

\begin{document}

\begin{frontmatter}

\title{MODL: A Modular Ontology Design Library}
\author{Cogan Shimizu, Quinn Hirt, \and Pascal Hitzler}
\institute{Data Semantics Laboratory, Wright State University, Dayton, OH, USA}

\maketitle

\begin{abstract}

Pattern-based, modular ontologies have several beneficial properties that lend themselves to FAIR data practices, especially as it pertains to Interoperability and Reusability. However, developing such ontologies has a high upfront cost, e.g. reusing a pattern is predicated upon being aware of its existence in the first place. Thus, to help overcome these barriers, we have developed MODL: a modular ontology design library. MODL is a curated collection of well-documented ontology design patterns, drawn from a wide variety of interdisciplinary use-cases. In this paper we present MODL as a resource, discuss its use, and provide some examples of its contents.

\end{abstract}

\end{frontmatter}

\section{Introduction}
\label{sec:intro}
The Information Age is an apt description for these modern times; between the World Wide Web and the Internet of Things an unfathomable amount of information is accessible to humans and machines, but the sheer volume and heterogeneity of the data have their drawbacks. Humans have difficulty drawing \emph{meaning} from large amounts of data. Machines can parse the data, but do not \emph{understand} it. Thus, in order to bridge this gap, data would need to be organized in such a way that some critical part of the human conceptualization is preserved. Ontologies are a natural fit for this role, as they may act as a vehicle for the sharing of \emph{understanding} \cite{gruber}.

Unfortunately, published ontologies have infrequently lived up to such a promise, hence the recent emphasis on FAIR (Findable, Accessible, Interoperable, and Reusable) data practices \cite{fair}. More specifically, many ontologies are not interoperable or reusable. This is usually due to incompatible ontological commitments: strong---or very weak---ontological committments lead to an ontology that is really only useful for a specific use-case, or to an ambiguous model that is almost meaningless by itself.

To combat this, we have developed a methodology for developing so-called modular ontologies \cite{chess}. In particular, we are especially interested in pattern-based modules \cite{momtut}. A modularized ontology is an ontology that individual users can easily adapt to their own use-cases, while still preserving relations with other versions of the ontology; that is, keeping it \emph{interoperable} with other ontologies. Such ontologies may be so adapted due to their ``plug-and-play'' nature; that is, one module may be swapped out for another developed from the same pattern.

An ontology design pattern is, essentially, a small self-contained ontology that addresses a general problem that has been observed to be invariant over different domains or applications \cite{odpbook}. By tailoring a pattern to a more specific use-case, an ontology engineer has developed a \emph{module}. This modelling paradigm moves much of the cost away from the formalization of a conceptualization (i.e. the logical axiomatization). Instead, pattern-based modular ontolody design (PBMOD) is predicated upon knowledge of available patterns, as well as being aware of the use-cases it addresses and its ontological commitments. 

Thus, in order to address the findability and accessibility aspects of PBMOD, we have developed MODL: a modular ontology design library. MODL is a curated collection of well-documented ontology design patterns. The particular research contribution is both the curation and documentation. Some of the patterns are novel, but many more have been extracted from existing ontologies and streamlined for use in a general manner. 
MODL, as an artefact, is distributed online as a collection of annotated OWL files and a technical report containing schema diagrams and explanations of each OWL axiom.\footnote{\url{https://dase.cs.wright.edu/content/modl-modular-ontology-design-library}} 

The rest of the paper is organized as follows. Section \ref{sec:rel} discusses the relevance of this work. Section \ref{sec:modl} presents our Modular Ontology Design Library in detail and in Section \ref{sec:conc} we conclude and discuss future work.

\section{Relevance}
\label{sec:rel}
This section provides an overview of where this resource fits into the state of the art, who is expected to use this resource, justification for why we believe that it will be used, and how we plan to disseminate the resource.
\subsection{Research Context}
\label{ssec:res}
Pattern-based modular ontology development is not a conceptually new idea---instead, it is a continuation of an already established paradigm. Both modularization of ontologies \cite{rector} and pattern-based modelling \cite{odp1} have been identified as improvements to the ontology engineering process. These concepts have culiminated in mature paradigms (e.g. MOM \cite{chess,momtut} and eXtreme Design \cite{extreme1,extreme2}), both having been used in large-scale projects (e.g. GeoLink \cite{geolink} and VALCRI \cite{valcri}). However, the ontology engineering community, especially those that utilize patterns, have indicated an increased need for better tooling support \cite{chap9,dagstuhlkg}, of which there are two complementary aspects: a dedicated development environment and a critical mass of Ontology Design Patterns (or ODPs for short).\footnote{Anecdotally, one of the more pervasive themes at both the 2018 and 2019 United States Semantic Technologies Symposia (\url{https://us2ts.org/}) was a call from ontology engineers in both academia and industry for better tooling support.}

There is already a prototype that begins to address the need for a dedicated development environment for pattern-based ontologies \cite{xd4p}. It also provides a set of hard-coded patterns that were extracted (at the time of development) from the ODP Portal.\footnote{\url{http://ontologydesignpatterns.org/wiki/Submissions:ContentOPs}} However, having the pattern library tightly coupled with the tool is disadvantageous for future development. Indeed, decoupling the tool is desirable, for a number of reasons, as follows. 
\begin{itemize}
\item Remove the onus of pattern development and upkeep off of the tool developer.
\item Enable community driven improvements and tailoring of the library to the end-users use-cases.
\item Enable plug-and-play pattern libraries for different domains, etc.
\end{itemize}
On the other hand, MODL also addresses the crucial need for a critical mass of ODPs. One may argue that this mass exists in the form of the ODP portal. Unfortunately, though, it has suffered under the weight of its own mission. Community enforced quality control has not succeeded in providing a ready-to-use suite of quality patterns for use across multiple domains. 

Furthermore, while the quality of a set of patterns is largely subjective, MODL strives for consistency in documentation, uses best practices \cite{how2doc,opla}, and limited ontological commitments. In some cases this required polishing extant documentation, writing it from scratch, and tweaking or detecting errors in the formalization. We also include all new schema diagrams \cite{karima} following a single paradigm and style.

MODL therefore addresses, in some fashion, both aspects of improving tooling support. In turn, we expect this to lower the barrier of entry to PBMOD, which in turn lowers the barrier of entry for wider adoption semantic web technologies in application areas.

\subsection{Adoption \& Dissemination}
\label{ssec:adopt}
As a new resource, we are certainly interested in reaching a wide audience, ensuring that the resource is visible to the community and adopted into existing workflows. In previous sections we have made the case that MODL is very much a response to a community need and that there is a vibrant, existing community surrounding modular ontology modelling and pattern-based modelling. Additionally, we are currently working very closely with the developer for \cite{xd4p} in order to integrate the MODL modality, i.e. ensure plug-and-play pattern libraries are supported by the dedicated development environment. We discuss concrete next steps in Section \ref{sec:conc}.

Additionally, we believe that MODL will help bridge communities that utilize a different conceptualization of patterns. For example, the pattern communities surrounding the biological ontologies (e.g. Gene Ontology \cite{gene} and the International Conference and Biological Ontology\footnote{\url{http://icbo2018.cgrb.oregonstate.edu/node/29}}), those communities that utilize OTTR \cite{ottr}, and the corresponding frame semantics found in Framester \cite{framester}. At a recent session\footnote{\url{http://us2ts.org/2019/posts/program-session-vii.html}} held during US2TS 2019, bridging the different design pattern communities was discussed; namely, how can future tools incorporate the lessons learned from these communities? These next steps are discussed in Sections \ref{ssec:meth} and \ref{sec:conc}.

Of course, there are the adjacent communities found in the Association for Ontology Design Patterns (ODPA)\footnote{\url{http://ontologydesignpatterns.org/wiki/ODPA}} and the long-standing Workshop on Ontology Design and Patterns,\footnote{\url{http://ontologydesignpatterns.org/wiki/WOP:Main}} where inspiration for this work originated. Thus, we believe it will be well received. We also wish to continue improving our tutorial on modular ontology modelling, where we think that MODL will be an excellent addition.\footnote{\url{http://ontologydesignpatterns.org/wiki/Training:Tutorial:_Methods_and_Tools_for_Modular_Ontology_Modeling}}

Finally, the authors have a number of ongoing collaborations with domain experts in the digital humanities. In terms of adoption and dissemination, this is perhaps our most important connection. Specifically, we are working on building an ontology for representing the records of enslaved people in the historic slave trade.\footnote{\url{https://enslave.org/}} A technical report detailing the modular ontology and its patterns can be found online \cite{enslaved}. There is a lot of excitement surrounding its potential for tying in a myriad of historians' work. Between MODL and \cite{xd4p}, we hope to enable an entire community to collaboratively building their ontologies. 

\section{A Modular Ontology Design Library}
\label{sec:modl}
In this section, we present in detail MODL. Section \ref{ssec:meth} explains our methodology and the organization of MODL, Section \ref{ssec:how} provides a brief tutorial on using MODL, Section \ref{ssec:exerpt} provides an example pattern that has been excerpted from the documentation (some of the language and structure, e.g. subsections, have been adapted to fit this paper format), and finally, Section \ref{ssec:deets} provides information pertaining to accessibility, sustainability, and more.

\subsection{MODL's Methodology}
\label{ssec:meth}
\begin{table}[t]
\begin{center}
\setlength\tabcolsep{0.3em}
\begin{tabular}{l | l}
\textbf{Category} & \textbf{Patterns}\\
\hline
\multirow{4}{41mm}{Metapatterns}
& Explicit Typing \\
& Property Reification \\
& Stubs \\
\hline
\multirow{4}{41mm}{Organization of Data}
& Aggregation, Bag, Collection \\
& Sequence, List\\ 
& Tree \\
\hline
\multirow{3}{41mm}{Space, Time, and Movement}
& Spatiotemporal Extent \\
& Spatial Extent \\
& Temporal Extent \\
& Trajectory \\
& Event \\
\hline 
\multirow{3}{41mm}{Agents and Roles}
& AgentRole \\
& ParticipantRole \\
& Name Stub \\
\hline
\multirow{4}{41mm}{Description and Details}
& Quantities and Units \\
& Partonymy/Meronymy \\
& Provenance \\
& Identifier \\
\end{tabular}
\caption{This table contains the patterns included in MODL. They have been partitioned into five categories (metapatterns; organization of data; space, time, and movement; agents and roles; and description and details) which are loosely defined by their general use-cases.}
\label{tab:pats}
\end{center}
\end{table}
MODL is a curated collection of well-documented ontology design patterns. MODL, itself, can be considered to be the combination of two artifacts, the collection of patterns, specified in OWL, and the accompanying documentation. The separation is a little fuzzy, as the OWL serialization is also heavily annotated for convenience. The mission of MODL is to make patterns both findable and accessible. Therefore, it is of utmost importance that every pattern therein is thoroughly documented. One drawback of the ODP Portal is that there are no guidelines provided for documenting the patterns and, during submission, a form is provided with many optional, ill-defined fields. That is not to say all of the patterns documented therein are poorly documented---some patterns did indeed have thorough documentation. Where possible, we preserved these efforts, from either the portal or associated publication, and corresponding credit is given in the MODL documentation.

However, for many of the patterns included in MODL, we needed to fill some gaps. For this we have elected to follow the guidelines set forth in \cite{how2doc}. These guidelines are a result of a community wide survey that ranks the perceived importance of ten different components of ODP documentation. For our purposes, we have chosen to include the top seven. They are \emph{Schema Diagram}, \emph{Example of Pattern Instantiation}, \emph{Compentency Questions}, \emph{Axiomatization}, \emph{OWL File}, \emph{Pointers to Related Patterns}, and \emph{Metadata}. The remaining three components (\emph{Set of Example SPARQL Queries}, \emph{Examples of Available Datasets for Population}, and \emph{Constraints Using ShEx}\footnote{\url{http://shex.io/}}) are being considered for future versions of MODL.\footnote{Furthermore, there is some community indecision on embracing ShEx or SHACL, a newer W3C recommendation. More information can be found at \url{https://www.w3.org/TR/shacl/}.}

The schema diagrams for our documentation were manually created using the algorithm found in \cite{karima,sdont}. We elected to use a simplified visual syntax that conveyed relations between concepts and also contains visual cues for identifying concepts that should be used as `hooks' into the ODP.

The provided OWL files for each of the patterns are annotated with the Extended Ontology Design Pattern Representation Language (OPLa)\footnote{\url{https://github.com/cogan-shimizu-wsu/Extended-OPLa}} \cite{opla}. This allows us to embed provenance metadata (e.g. where did this pattern originate?) or provide pointers to related patterns (e.g. generalizations or specializations of the pattern) in annotations. We discuss future extensions to OPLa and how they may be leveraged to improve MODL in Section \ref{sec:conc}.

Table \ref{tab:pats} lists the patterns included in MODL. They have been loosely organized into five categories: metapatterns; organization of data; space, time, and movement; agents and roles; and description and details. 

\noindent\textbf{Metapatterns} This category contains patterns that can be considered to be ``patterns for patterns.'' In other literature, notably \cite{odp1}, they may be called \emph{structural ontology design patterns}, as they are independent of any specific context, i.e. they are content-independent. This is particularly true for the metapattern for property reification, which, while a modelling strategy, is also a workaround for the lack of $n$-ary relationships in OWL. The other metapatterns address structural design choices frequently encountered when working with domain experts. They present a best practice to non-ontologists for addressing language specific limitations.

\noindent\textbf{Organization of Data} This category contains patterns that pertain to how data might be organized. These patterns are necessarily highly abstract, as they are ontological reflections of common data structures in computer science. The pattern for aggregation, bag, or collection is a simple model for connecting many concepts to a single concept. Analogously, for the list and tree pattterns, which aim to capture ordinality and acyclicity, as well. More so than other patterns in this library, these patterns provide an axiomatization as a high-level framework that must be specialized (or modularized) to be truly useful.

\noindent\textbf{Space, Time, and Movement} This category contains patterns that model the movement of a thing through a space or spaces and a general event pattern. The semantic trajectory pattern is a more general pattern for modelling the discrete movements along some dimensions. The spatiotemporal extent pattern is a trajectory along the familiar dimensions of time and space. Both patterns are included for convenience.

\noindent\textbf{Agents and Roles} This category contains patterns that pertain to agents interacting with things. Here, we consider an agent to be anything that performs some action or role. This is important, as it decouples the role of an agent from the agent itself. For example, a \textsf{Person} may be \textsf{Husband} and \textsf{Widower} at some point, but should not be both simultaneously. These patterns enable the capture of this data. In fact, the agent role and participante role patterns are convenient specializations of property reification that have evolved into a modelling practice writ large. In this category, we also include the name stub, which is a convenient instantiation of the stub metapattern; it allows us to acknowledge that a name is a complicated thing, but sometimes we only really need the string representation.

\noindent\textbf{Description and Details} This category contains patterns that model the description of things. These patterns are relatively straightforward, models for capturing ``how much?'' and ``what kind?'' for a particular thing; patterns that are derived from Winston's part-whole taxonomy \cite{partof}; a pattern extracted from PROV-O \cite{provo}, perhaps to be used to answer ``where did this data come from?''; and a pattern for associating an identifier with something.

\subsection{Using MODL}
\label{ssec:how}
There are two different ways to use MODL---for use in ontology modelling and for use in tools. In both cases, MODL is distributed as a ZIP archive of the patterns' OWL files and accompanying documentation. In the case of the Ontology Engineer, it is simply used as a resource while building an ontology, perhaps by using Modular Ontology Modelling or eXtreme Design methodologies. For the tool developer, we also supply an ontology consisting of exactly the OPLa annotations from each pattern that pertain to \textsf{OntologicalCollection}. As OPLa is fully specified in OWL, these annotations make up an ontology of patterns and their relations. One particular use-case that we foresee is a tool developer querying the ontology for which patterns are related to the current pattern, or looking for a pattern based on keywords or similarity to competency questions.

\subsection{Excerpt from Pattern Documentation}
\label{ssec:exerpt}

\bigskip
\noindent\textbf{EntityWithProvenance}
\begin{figure}[h!]
\begin{center}
\includegraphics[width=.5\textwidth]{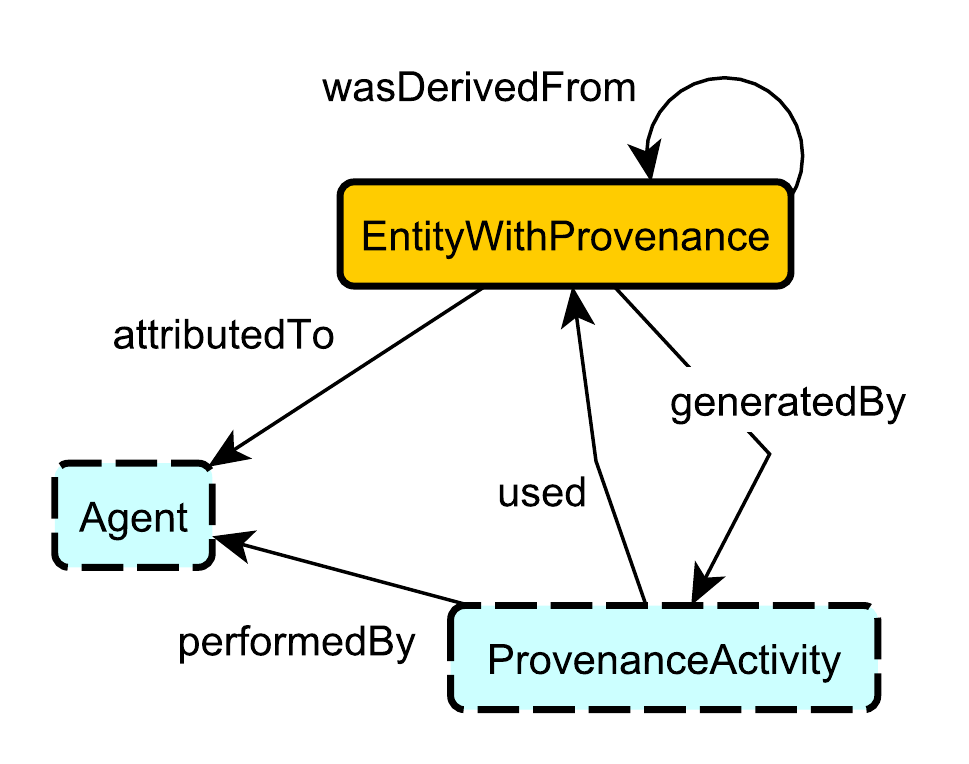}
\caption{This figure depicts the schema diagram for the \textsf{EntityWithProvenance} Pattern which is essentially the core of the Provenance Ontology (PROV-O). Yellow boxes are concepts. Light blue boxes with a dashed border are external patterns that the developer may want to also make into a module.}
\label{fig:prov}
\end{center}
\end{figure}

\noindent\textbf{Summary}\\
\noindent Figure \ref{fig:prov} depicts the schema diagram for the Provenance pattern, as included in MODL. The \textsf{EntityWithProvenance} Pattern is extracted from the PROV-O ontology. At the pattern level, we do not want to make the ontological committment to a full-blown ontology. It suffices to align a sub-pattern to the core of PROV-O. \cite{provo}

The \textsf{EntityWithProvenance} class is any item of interest to which a developer would like to attach provenance information. That is they are interested in capturing, who or what created that item, what was used to derive it, and what method was used to do so. The ``who or what'' is captured by using the \textsf{Agent} class. The property, \textsf{wasDerivedFrom} is eponymous---it denotes that some set of resources was used during the \textsf{ProvenanceActivity} to generate the \textsf{EntityWithProvenance}.

\bigskip

\noindent\textbf{Axiomatization}\footnote{Axiomatization is extensive while avoiding undesirably strong ontological commitments. Most axioms for the MODL patterns follow the template of the OWLAx Prot\'eg\'e plug-in \cite{SarkerKH16}.}
\begin{align}
\exists\textsf{attributedTo}.\textsf{Agent} &\sqsubseteq \textsf{EntityWithProvenance}\\
\textsf{EntityWithProvenance} &\sqsubseteq \forall\textsf{attributedTo}.\textsf{Agent}\\
\exists\textsf{generatedBy}.\textsf{ProvenanceActivity} &\sqsubseteq \textsf{EntityWithProvenance}\\
\textsf{EntityWithProvenance} &\sqsubseteq \forall\textsf{generatedBy}.\textsf{ProvenanceActivity}\\
\exists\textsf{used}.\textsf{EntityWithProvenance} &\sqsubseteq \textsf{ProvenanceActivity}\\
\textsf{ProvenanceActivity} &\sqsubseteq \forall\textsf{used}.\textsf{EntityWithProvenance}\\
\exists\textsf{performedBy}.\textsf{Agent} &\sqsubseteq \textsf{ProvenanceActivity}\\
\textsf{ProvenanceActivity} &\sqsubseteq \forall\textsf{performedBy}.\textsf{Agent}
\end{align}
\noindent\textbf{Axiom Explanations}
\begin{enumerate}
\item Scoped Domain:The scoped domain of \textsf{attributedTo}, scoped by \textsf{Agent}, is \textsf{EntityWithProvenance}.
\item Scoped Range: The scoped range of \textsf{attributedTo}, scoped by \textsf{EntityWithProvenance}, is \textsf{Agent}. 
\item Scoped Domain:The scoped domain of \textsf{generatedBy}, scoped by \textsf{ProvenanceActivity}, is \textsf{EntityWithProvenance}.
\item Scoped Range: The scoped range of \textsf{generatedBy}, scoped by \textsf{EntityWithProvenance}, is \textsf{ProvenanceActivity}.
\item Scoped Domain:The scoped domain of \textsf{used}, scoped by \textsf{EntityWithProvenance}, is \textsf{ProvenanceActivity}
\item Scoped Range: The scoped range of \textsf{used}, scoped by \textsf{ProvenananceActivity}, is \textsf{EntityWithProvenance}.
\item Scoped Domain:The scoped domain of \textsf{performedBy}, scoped by \textsf{Agent}, is \textsf{ProvenanceActivity}.
\item Scoped Range: The scoped range of \textsf{performedBy}, scoped by \textsf{ProvenanceActivity}, is \textsf{Agent}.
\end{enumerate}
\textbf{Competency Questions}
\begin{enumerate}[CQ1.]
\item Who are the contributors to this Wikidata page?
\item From which database is this entry taken?
\item Which method was used to generate this chart and from which spreadsheet did the data originate?
\item Who provided this research result?
\end{enumerate}

\subsection{Details}
\label{ssec:deets}
\paragraph{Persistent URI} The persistent URI for this resource is \url{https://archive.org/services/purl/purl/modular_ontology_design_library}. The Version 1.0 snapshot and its documentation may be found there. Additionally, it provides helpful links to a technical report and the living data on GitHub, as discussed below.

\paragraph{Canonical Citation} The canonical citation for this resource may be found on arXiv \cite{canon}.

\paragraph{Documentation} In addition to this document, we provide in-depth documentation on the library. This documenation contains a primer on ontology design patterns, as a concept, as well as common techniques used in their formalization. Most importantly, for each pattern it provides a schema diagram, its axiomatization, and explanations for each of those axioms. As mentioned in Section \ref{ssec:meth}, each pattern is thoroughly annotated with OPLa which provides further documentation on its use and provenance.

\paragraph{Sustainability \& Maintenance} MODL straddles the realms of dataset and software library; the resource is essentially a snapshot of data that lives. Due to this potential for change, we intend to maintain MODL analogously to a software project. Indeed, while the snapshots will be distributed as ZIP archives, the living data is (at the time of this writing) hosted on GitHub.\footnote{\url{https://github.com/cogan-shimizu-wsu/modular-ontology-design-library}} The Data Semantics Laboratory\footnote{\url{http://daselab.org/}} will host MODL's snapshots and appropriate documentation indefinitely. The authors plan to drive further development of needed or requested patterns. Furthermore, by using Git\footnote{\url{https://git-scm.com/}} we inherit mechanisms for tracking issues and versions and incorporating such community contributions into future releases. 

\paragraph{License Information} This resource is released under the Creative Commons Attribution 4.0 International Public License the details of which can be found online.\footnote{\url{https://creativecommons.org/licenses/by/4.0/legalcode}}
\section{Conclusions}
\label{sec:conc}
MODL is a curated collection of well-documented ontology design patterns. We have created this resource to meet a community-recognized need for tooling infrastructure for ontology engineering. In particular, this resource makes ontology design patterns both findable and accessible, shows how they are interoperable, and promotes their reuse. Furthermore, we posit that future ontologies reusing these patterns will promote their interoperability and reuse.

\subsection{Next Steps}
The next steps are many, as MODL is a foundational resource. We have identified several patterns that we deem necessary for covering additional frequently encountered modelling needs, e.g. a process pattern or patterns. As mentioned in Section \ref{ssec:meth}, we also want to further flesh out the documentation with respect to \cite{how2doc}. One future use case that we foresee for this resource is the mapping of competency questions to example SPARQL queries, which maybe could be used as a gold-standard training set for an automated translator. Also mentioned in Section \ref{ssec:meth}, we intend to work closely with the digital humanities community for their knowledge representation needs. Finally, we have noted the extreme importance of working closely with tool developers; there is ongoing work with the developer of \cite{xd4p} to create a Prot\'eg\'e plug-in that utilizes MODL as a base for modular ontology modelling.


\medskip

\noindent\emph{Acknowledgement.} Cogan Shimizu acknowledges support by the Dayton Area Graduate Studies Institute (DAGSI). Quinn Hirt acknowledges funding from the Air Force Office of Scientific Research under award number FA9550-18-1-0386.

\bibliographystyle{abbrv}
\bibliography{refs.bib}

\end{document}